\documentclass[10pt,letterpaper]{article}
\usepackage[top=0.85in,left=2.75in,footskip=0.75in,marginparwidth=2in]{geometry}

\usepackage[utf8]{inputenc}

\usepackage{cite}

\usepackage{nameref,hyperref}

\usepackage[right]{lineno}

\usepackage{microtype}
\DisableLigatures[f]{encoding = *, family = * }

\raggedright
\usepackage{changepage}

\usepackage[aboveskip=1pt,labelfont=bf,labelsep=period,singlelinecheck=off]{caption}

\makeatletter
\renewcommand{\@biblabel}[1]{\quad#1.}
\makeatother

\usepackage{lastpage,fancyhdr,graphicx}
\usepackage{epstopdf}
\fancyheadoffset[L]{2.25in}
\fancyfootoffset[L]{2.25in}

\usepackage{color}

\definecolor{Gray}{gray}{.25}

\usepackage{graphicx}

\usepackage{sidecap}

\usepackage{amsmath,amssymb}
\DeclareMathOperator{\E}{\mathbb{E}}

\usepackage{wrapfig}
\usepackage[pscoord]{eso-pic}
\usepackage[fulladjust]{marginnote}
\reversemarginpar

\usepackage[nodayofweek]{datetime}
\newdateformat{mytoday}{\monthname[\THEMONTH] \twodigit{\THEDAY}, \THEYEAR}

\usepackage{multirow}
\usepackage{booktabs}
\usepackage{rotating}
\usepackage{tabulary}
\newcolumntype{K}[1]{>{\centering\arraybackslash}p{#1}}

\begin{document}
\noindent Cite this preprint version of the manuscript as:

\ \\
\noindent \textcolor{blue}{M. Mahmud, M.S. Kaiser, A. Hussain, S. Vassanelli. ``Applications of Deep Learning and Reinforcement Learning to Biological Data,'' \textit{IEEE Trans. Neural Netw. Learn. Syst.}, 2018, doi: 10.1109/TNNLS.2018.2790388.
\ \\
\noindent \textcopyright\ IEEE holds the copyright of this work.
}

\vspace*{0.35in}
\begin{flushleft}
{\Large
\textbf\newline{Applications of Deep Learning and Reinforcement Learning to Biological Data}
}
\newline
\\
Mufti Mahmud\textsuperscript{1,*},
M. Shamim Kaiser\textsuperscript{2,*},
Amir Hussain\textsuperscript{3},
Stefano Vassanelli\textsuperscript{1}
\\
\bigskip
\textsuperscript{1} NeuroChip Lab, University of Padova, 35131 - Padova, Italy
\\
\textsuperscript{2} IIT, Jahangirnagar University, Savar, 1342 -  Dhaka, Bangladesh
\\
\textsuperscript{3} Division of Computing Science \& Maths, University of Stirling, FK9 4LA Stirling, UK \\
\bigskip
\textsuperscript{*} Co-`first and corresponding' author. 
Emails: muftimahmud@gmail.com (M. Mahmud), mskaiser@juniv.edu (M.S. Kaiser) \\
\bigskip

\end{flushleft}

\section*{Abstract}

Rapid advances of hardware-based technologies during the past decades have opened up new possibilities for Life scientists to gather multimodal data in various application domains (e.g., \textit{Omics}, \textit{Bioimaging}, \textit{Medical Imaging}, and \textit{[Brain/Body]-Machine Interfaces}), thus generating novel opportunities for development of dedicated data intensive machine learning techniques.
Overall, recent research in Deep learning (DL), Reinforcement learning (RL), and their combination (Deep RL) promise to revolutionize Artificial Intelligence. The growth in computational power accompanied by faster and increased data storage and declining computing costs have already allowed scientists in various fields to apply these techniques on datasets that were previously intractable for their size and complexity. This review article provides a comprehensive survey on the application of DL, RL, and Deep RL techniques in mining Biological data. In addition, we compare performances of DL techniques when applied to different datasets across various application domains. Finally, we outline open issues in this challenging research area and discuss future development perspectives.



\section*{Introduction}
\label{sec-intro}

The need for novel healthcare solutions and continuous efforts in understating the biological bases of pathologies have pushed extensive research in the Biological Sciences over the last two centuries \cite{Coleman_biology_1977}. Recent technological advancements in Life Sciences opened up possibilities not only to study Biological systems from a holistic perspective but provided unprecedented access to the molecular details of the living organisms \cite{Magner_history_2002,Brenner_history_2012}.
Novel tools for DNA sequencing \cite{shendure_next-generation_2008}, gene expression \cite{metzker_sequencing_2010}, bioimaging \cite{Vadivambal_bioimaging_2016}, neuroimaging \cite{poldrack_progress_2015}, and brain-machine interfaces \cite{Lebedev-bmi-2017} are now available to the scientific community. However, considering the inherent complexity of the biological systems together with the high-dimensionality, diversity, and noise contaminations, inferring meaningful conclusion from these data is a huge challenge \cite{marx_biology_2013}. Therefore, novel instruments are required to process and analyze biological big data that must be robust, reliable, reusable, and accurate \cite{li_big_2014}. 
This encouraged numerous scientists from life and computing sciences disciplines to embark in a multidisciplinary approach to demystify functions and dynamics of living organisms with remarkable progress to biological and biomedical research \cite{wickware_next-generation_2000}.
Thus, many techniques of Artificial Intelligence (AI), in particular machine learning (ML), have been proposed over time to facilitate recognition, classification, and prediction of patterns in biological data \cite{tarca_machine_2007}.

The conventional ML techniques can be broadly categorized in two large sets -- \textit{supervised} and \textit{unsupervised}. The methods pertaining to the \textit{supervised} learning paradigm classify objects in a pool using a set of known annotations/ attributes/ features.
Instead, the \textit{unsupervised} learning techniques form groups/ clusters among the objects in a pool by identifying their similarity and then use them for classifying the unknowns.
Also, the other category, \textit{reinforcement learning} (RL), allows a system to learn from the experiences it gains through interacting with its environment (see section \ref{subsec-overview-rl} for details).

Popular \textit{supervised} methods include: Artificial Neural Network (ANN) \cite{hopfield_artificial_1988} and its variants, 
Support Vector Machines \cite{cortes_support-vector_1995} and linear classifiers \cite{yuan_recent_LC_2012}, 
Bayesian Statistics \cite{heckerman_tutorial_1998},  
k-Nearest Neighbors \cite{cover_knn_1967}, Hidden Markov Model \cite{rabiner_hmm_1986}, and Decision Trees \cite{kohavi_dt_2002}. Also, popular \textit{unsupervised} methods include: Autoencoders \cite{Hinton-autoencoder-1989}, Expectation Maximization \cite{dempster_em_1977}, 
Self-Organizing Maps \cite{kohonen_som_1982}, 
k-Means \cite{ball_km_1965}, and Fuzzy \cite{dunn_fc_1973} and Density-based \cite{hartigan_dbc_1975} clustering.

\begin{wrapfigure}[30]{l}{76mm}
\includegraphics[width=75mm]{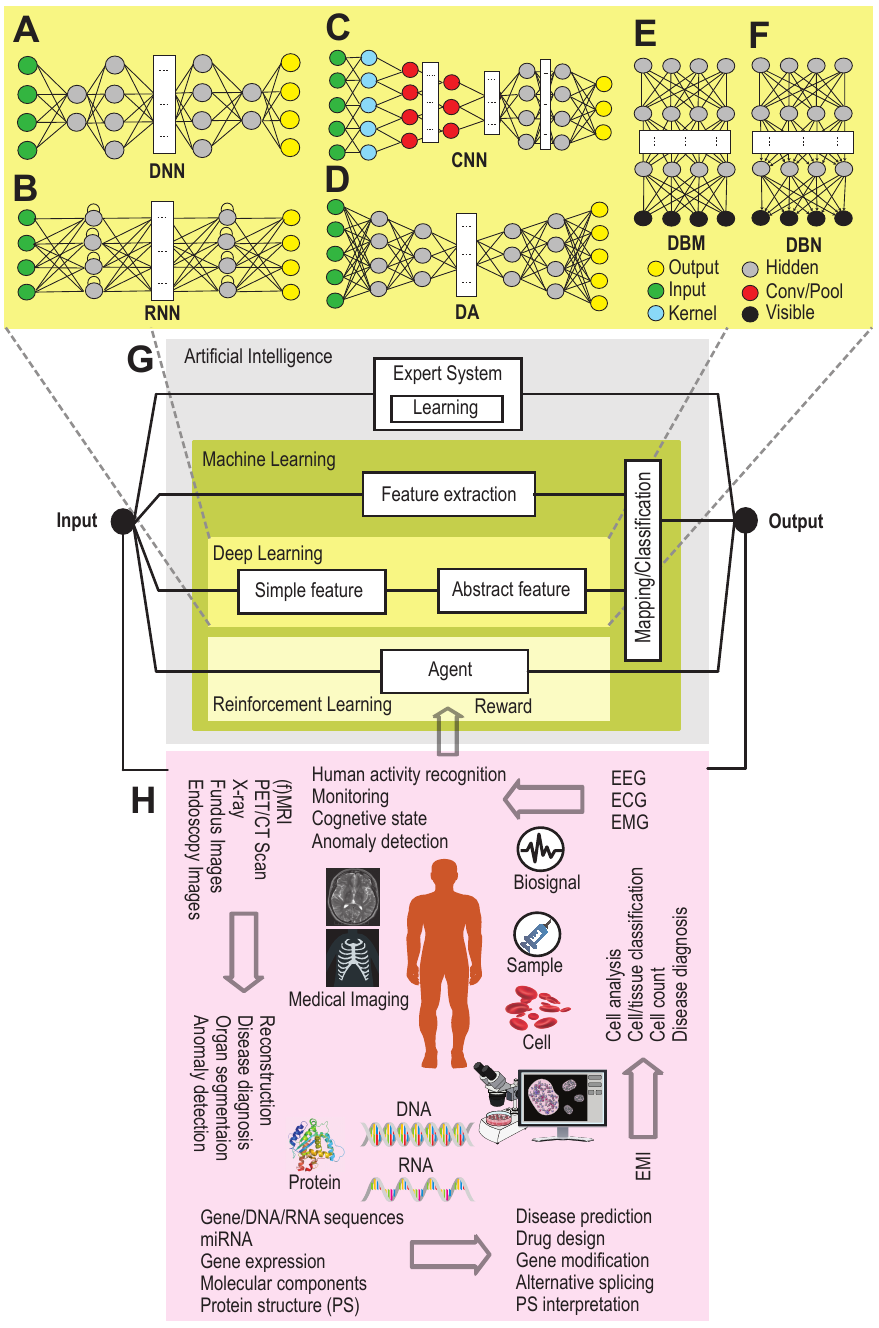}
\captionsetup{labelformat=empty} 
\caption{}
\label{fig_1} 
\end{wrapfigure}

\marginpar{
\vspace{.1cm} 
\textbf{Figure \ref{fig_1}} 
A possible representation of the DL, RL, and deep RL frameworks for biological applications. \textbf{A--F}. The popular DL architectures. \textbf{G}. Schematic diagram of the learning framework as a part of Artificial Intelligence (AI). Broadly, AI can be thought to have evolved parallelly in two main directions-- Expert Systems (ES) and ML. ES takes expert decisions from given factual data using rule based inferences. ML extracts features from data mainly through statistical modeling and provides predictive output when applied to unknown data. DL, being a sub-division of ML, extracts more abstract features from a larger set of training data mostly in a hierarchical fashion resembling the working principle of our brain. 
The other sub-division, RL, provides a software agent which gathers experience based on interactions with the environment through some actions and aims to maximize the cumulative performance. 
\textbf{H}. Possible applications of AI to biological data.
}

A large body of evidences shows that the above mentioned methods and their respective variants can be successfully applied to Biological data coming from various sources,
e.g., \textit{Omics} (covers data from genetics and [gen/ transcript/ epigen/ prote/ metabol]omics 
\cite{libbrecht_ml_2015}), 
\textit{Bioimaging} (covers data from [sub-]cellular images acquired by diverse imaging techniques \cite{kan_machine_2017}), \textit{Medical Imaging} (covers data from [medical/ clinical/ health] imaging mainly through diagnostic imaging techniques \cite{erickson_machine_2017}), 
and \textit{[Brain/Body]-Machine Interfaces or BMI} (covers electrical signals generated by the Brain and the Muscles and acquired using appropriate sensors \cite{vidaurre_machine-learning-based_2010,mahmud_processing_2016}).

Broadly, AI can be thought to have evolved parallelly in two main directions-- Expert Systems and ML (see the schematic diagram of Fig. \ref{fig_1}H). 
Focusing on the latter, ML extracts features from  training dataset(s) and make models with minimal or no human intervention. These models provide predicted outputs based on test data. DL, being a sub-division of ML, extracts more abstract features from a larger set of training data mostly without human supervision.
RL, being the other sub-division of ML, is inspired by psychology. It provides a software agent which gathers experience based on interactions with the environment through some actions and aims to maximize the cumulative performance. 

In recent years DL, RL, and deep RL methods are poised  to reshape the future of ML 
\cite{mnih_human-level_2015}. 
Over the last decade, the works pertaining to DL, RL, and deep RL were extensively reviewed from different perspectives. In a topical review, Schmidhuber provided a detailed time line of significant DL developments (for both supervised and unsupervised), RL and evolutionary computation, and DL in feed-forward and recurrent neural networks (NNs) for RL \cite{schmidhuber_deep_2015}. Other reviews are focusing on applications of DL in health informatics \cite{ravi_dl_2017}, biomedicine \cite{mamoshina_applications_2016}, and bioinformatics \cite{min_dl_bio_2016}. On the other hand, Kaelbling et al. discussed RL from the perspective of the trade-off between exploitation and exploration, RL's foundation via Markov decision theory, the learning mechanism using delayed reinforcement, construction of empirical learning models, use of generalization and hierarchy, and reported some exemplifying RL systems implementations \cite{kaelbling_rl_1996}. Glorennec provides a brief overview of the basis of RL with explicit descriptions of Q- and Fuzzy Q-learning \cite{glorennec_rl_review_2000}. With respect  to applications in solving dynamic optimization problems, Gosavi surveys Q-learning, temporal differences, semi-Markov decision problems, stochastic games, policy gradients, and hierarchical RL with detailed underlying mathematics \cite{gosavi_rl_review_2009}. In addition, Li analyzed the recent advances of deep RL on-- Deep Q-Network (DQN) with its extensions, asynchronous methods, policy optimization, reward, and planning as well as different applications including games (e.g., AlphaGo, robotics, chatbot, etc.), neural architecture design, natural language processing, personalized web services, healthcare, and finance \cite{li_deeprl_review_2017}.

Despite the popularity of the topic and application potential to diverse disciplines, a comprehensive review is missing  that focuses on data from different Biological application domains while providing a performance comparison across techniques. This review is intended to fill this gap: it provides a brief overview on DL, RL, and deep RL concepts, followed by state-of-the-art applications of these techniques and performance comparison between various DL approaches. Finally, it 
identifies and outlines some open issues and speculates about future perspectives.

As for the organization of the rest of the article, section \ref{sec-overview} provides a conceptual overview to the DL, RL, and deep RL techniques, thus introducing the reader to the underlying theory; section \ref{sec-applications} contains the state-of-the-art applications of these techniques to various biological application domains; 
section \ref{sec-perf-comp} presents  test results and performance comparison of DL techniques applied on datasets pertaining to different application domains;
section \ref{sec-issues-persp} highlights open issues and hints on future perspectives; and is concluded in section \ref{sec-conclusion}.


\section{Conceptual Overview}
\label{sec-overview}



\subsection{Deep Learning}
\label{subsec-overview-dl}

The core concept of DL is to learn data representations through increasing abstraction levels. Almost in all levels more abstract representations at a higher level are learned by defining them in terms of less abstract representations at lower levels. This type of hierarchical learning process is very powerful as it allows a system to comprehend and learn complex representations directly from the raw data \cite{bengio_learning_2009}, making it useful in many disciplines \cite{Goodfellow-et-al-2016}. 

Several DL architectures have been reported in the literature including: Deep Neural Network (DNN), Recurrent Neural Network (RNN), Convolutional Neural Network (CNN), Deep Autoencoder (DA), Deep Boltzmann Machine (DBM), Deep Belief Network (DBN), Deep Residual Network, Deep Convolutional Inverse Graphics Network, etc. For the sake of brevity, only the ones widely used with Biological data are  briefly summarized below. However, the interested readers are redirected to the references mentioned in each subsection for concrete mathematical details behind each architecture.

\subsubsection{Deep Neural Network}
\label{subsubsec-theo-DNN}
DNN (Fig. \ref{fig_1} A) \cite{saxe_dlnn_2013} is inspired by the brain's visual input processing mechanism which takes place at multiple levels (i.e., starting with cortical area `V1' and then passing to area `V2', and so on) \cite{schmidhuber_deep_2015}. The standard neural network (NN) is extended to have multiple hidden layers with nonlinear modules embodied in each hidden layer allowing it to learn part-hole of the representations. Though this formulation has been successfully used in many applications, the training process is slow and cumbersome.

\subsubsection{Recurrent Neural Network}
\label{subsubsec-theo-RNN}
RNN (Fig. \ref{fig_1} B) \cite{pascanu_drnn_2014} is a NN model designed to detect structures in streams of data \cite{elman_finding_1990}. Unlike feedforward NN which performs computations unidirectionally from input to output, RNN computes the current state's output depending on the outputs of the previous states. Due to this `memory'-like property, despite learning problems related to vanishing and exploding gradients, RNN gained popularity in many fields involving streaming data (e.g., text mining, time series, genomes, etc.). In recent years, two main variants, bidirectional RNN (BRNN) \cite{schuster_bidirectional_rnn_1997} and long short-term memory (LSTM) \cite{hochreiter_lstm_1997} have also been applied \cite{lipton_critical_2015,Lecun2015}.

\subsubsection{Convolutional Neural Network}
\label{subsubsec-cnn}
CNN (Fig. \ref{fig_1} C) \cite{wiatowski_dcnn_2015} is a multilayer NN model \cite{lecun_cnn_1998}, inspired by the neurobiology of visual cortex, that consists of convolutional layer(s) followed by fully connected layer(s). In between these two types of layers the may exist subsampling steps. They get the better of DNNs which have difficulty in scaling well with multidimensional locally correlated input data. Therefore, the main application of CNN has been in datasets where the number of nodes and parameters required to be trained is relatively large (e.g., image analysis). Exploiting the `stationary' property of an image, convolution filters (CF) can learn data-driven kernels. Applying such CF along with a suitable pooling function reduces the features that are supplied to the fully connected network to  classify. However, in case of large datasets even this can be daunting and can be solved using sparsely connected networks. Some of the popular CNN configurations include: AlexNet \cite{krizhevsky_alexnet_2012}, VGGNet \cite{simonyan_vgg_2014}, and GoogLeNet \cite{szegedy_googlenet_2015}.

\subsubsection{Deep Autoencoder}
\label{subsubsec-theo-DA}
DA architecture (Fig. \ref{fig_1} D) \cite{vincent_dae_2010} is obtained by stacking a number of Autoencoders which are data driven NN models (i.e., unsupervised) designed to reduce data dimension by automatically projecting incoming representations to a lesser dimensional space than that of the input. 
In an Autoencoder, equal amount of units are used in the input/output layers and less units in the hidden layers. (Non)linear transformations are embodied in the hidden layer units to encode the given input into smaller dimensions \cite{baldi_autoencoder_2012}. Despite that it requires a pre-training stage and suffers from vanishing error, this architecture is popular for its data compression capability and have many variants, e.g., Denoising Autoencoder \cite{vincent_dae_2010}, Sparse Autoencoder \cite{ranzato_sae_2006}, Variational Autoencoder \cite{kingma_vae_2014}, and Contractive Autoencoder \cite{rifai_cae_2011}.

\subsubsection{[Restricted] Boltzmann Machine ([R]BM)}
\label{subsubsec-theo-RBM}
[R]BM is an undirected probabilistic generative model representing specific probability distributions \cite{Salakhutdinov_dbm_2009}. It is also considered as nonlinear feature detector. The learning process of [R]BM is based on optimizing its parameters for a set of given observations to obtain the best possible fit of the probability distribution through Gibbs sampling (a Markov Chain Monte Carlo method \cite{geman_gibbs_sampling_1984})  \cite{fischer_rbm_2012}. BM has symmetrical connections among its units and has one visible layer with (multiple) hidden layers. Usually, the learning process of a BM is slow and computationally expensive, thus, requires long to reach equilibrium statistics \cite{bengio_learning_2009}. By restricting the intralayer units of a BM to connect among themselves a bipartite graph is formed (i.e., an RBM has a visible and a hidden layer) where the learning inefficiency is solved \cite{Salakhutdinov_dbm_2009}. Stacking multiple RBMs as learning elements yields the following two DL architectures. 

\paragraph{Deep Boltzmann Machine}
\label{para-dbm}
DBM (Fig. \ref{fig_1} E) \cite{desjardins_training_dbm_2012} is a stack of undirected RBMs. Being undirected, there is a feedback process among the layers where feature inference from higher level units affect the inference of lower level units. Despite this powerful inference mechanism which allows an input's alternative interpretations through concurrent competition at all levels of the model, estimating model parameters from data remains difficult. Gradient based methods (e.g., persistent contrastive divergence \cite{tieleman_rbm_lg_2008}) fail to explore the model parameters sufficiently \cite{desjardins_training_dbm_2012}. Though this learning problem is overcome by pretraining each RBM in a layerwise greedy fashion, with outputs of the hidden variables from lower layers as input to upper layers \cite{Salakhutdinov_dbm_2009}, the time complexity remains high and may not be suitable for large training datasets \cite{guo_deep_2016}. 

\paragraph{Deep Belief Network}
\label{para-dbn}
DBN (Fig. \ref{fig_1} F) \cite{hinton_dbn_2006} is formed by ordering several RBMs in a way that one RBM's latent layer is linked to the subsequent RBM's visible layer. The connections of DBN are downward directed to its immediate lower layer, except that the upper two layers are undirected \cite{hinton_dbn_2006}. Thus, DBN is a hybrid model with the first two layers as undirected graphical model and the rest being directed generative model. The different layers are learned in a layerwise greedy fashion and fine tuned based on required output \cite{ravi_dl_2017}, however, the training procedure is computationally demanding. 

\subsection{Reinforcement Learning}
\label{subsec-overview-rl}

Rooted in behavioral psychology, RL is a distinctive member of the ML family. An RL problem is solved by learning new experiences through trial--and--error. An RL \textit{agent} is trained, as such, it's \textit{actions} to interact with the \textit{environment} maximizes the cumulative \textit{reward} resulting from the interactions. 
Generally, RL problems are modeled and solved using Markov Decision Processes (MDP) theory through Monte Carlo (MC) and dynamic programming (DP) \cite{sutton_rl_book_1998}.

The learning of an agent is a continuous process where the interactions with the environment occurs at discrete time steps. In a typical RL cycle (at time $t$), the agent receives the environment's state (i.e., \textit{state}, $s_t$) and selects an action ($a_{t}$) to interact. The environment responds to the action and progresses to a new state ($s_{t+1}$). The reward ($r_{t+1}$), that the agent either receives or not for the selected action, associated to the \textit{transition} ($s_t$, $a_t$, $s_{t+1}$) is also determined \cite{sutton_rl_book_1998}. Accordingly, after each cycle, the agent updates the \textit{value function} $V(s)$ or \textit{action-value function} $Q(s,a)$ based on certain \textit{policy}, where, \textit{policy} ($\pi$) is a function that maps states $s\in S$ to actions $a\in A$, i.e., $\pi: S \rightarrow A \Rightarrow a=\pi(s)$ \cite{kaelbling_rl_1996}. 

A possible way to solve the RL problem is to describe the environment as MDP with a set of state-value function pairs, a set of actions, a policy, and a reward function. The value function can be separated to solve state-value function ($V$) or action-value function ($Q$). In the state-value function the expected outcome, of being in state $s$ following policy $\pi$, is determined by sum of the rewards at future time steps with a given discount factor 
($\gamma \in [0, 1]$), i.e., $V^\pi(s) = \E_\pi (\sum_{k=0}^\infty \gamma^k r_{t+k+1} | s_t = s)$.
And in the action-value function the expected outcome, of being in state $s$ taking action $a$ following policy $\pi$, is determined by sum of the rewards for each state action pairs, i.e., 
$Q^\pi(s, a) = \E_\pi (\sum_{k=0}^\infty \gamma^k r_{t+k+1} | s_t = s, a_t=a)$.

The MDP can be solved and the optimum policy can be achieved through DP by: either starting with an initial policy and improving it iteratively (\textit{policy iteration}), or starting with arbitrary value function and recursively refining an estimate of an improved state-value or action-value function to compute an optimal policy and its value (\textit{value iteration}) \cite{poole_ai_2017}. In the simplest case,
the state-value function for a given policy can be estimated using Bellman expectation equation as:
$V^{\pi}(s)=\E_\pi(r_{t+1}+\gamma V^\pi(s_{t+1})|s_t=s)$.
Considering this as a \textit{policy evaluation} process, an improved and eventually optimal policy ($\pi^*$) can be achieved by taking actions greedily that maximizes the state-action value.
But in scenarios with unknown environments, model-free methods are to be used without MDP. In such cases, instead of the state-value function, the action-value function can be maximized to find the optimal policy ($\pi^*$) using a similar policy evaluation and improvement process, i.e., $Q^\pi(s,a)=\E_\pi(r_{t+1}+\gamma Q^\pi(s_{t+1},a_{t+1})|s_t=s,a_t=a)$. There are several learning techniques, e.g., Monte Carlo, Temporal Difference (TD), and State-Action-Reward-State-Action (SARSA), which describe various aspects of the model-free policy evaluation and improvement process \cite{busoniu_rl_2010}.

However, in real world RL problems, the state-action space is very large and storing a separate value function for every possible state is cumbersome. In such situations generalization of the value function through \textit{function approximation} is required. For example, the $Q$ value function approximation is able to generalize to unknown states by calculating a function ($\hat{Q}$) for a given state action pair ($s,a$), i.e., $\hat{Q}(s,a,\textbf{w}) \approx Q^\pi(s,a)=\textbf{x}(s,a)^\top \textbf{w}$. In other words, a rough approximation of the $Q$ function is obtained from the feature vector representing $(s,a)$ pair ($\textbf{x}$) and the provided parameter ($\textbf{w}$ which is updated using MC or TD learning) \cite{sutton_fast_2009}. This approximation allows to improve the $Q$ function by minimizing the loss between the true and approximated values (e.g., using gradient descent), i.e., $J(\textbf{w})=\E_\pi((Q_\pi(s,a)-\hat{Q}(s,a,\textbf{w}))^2)$. Examples of differentiable function approximators include: neural network, linear combinations of features, decision tree, nearest neighbor, Fourier bases, etc. \cite{schaul_universal_2015}.

\subsection{Deep Reinforcement Learning}
\label{subsec-theory-DRL}

The autonomic capability to learn without any feature crafting makes RL a powerful tool applicable to many disciplines, but it falls short in cases when the data dimensionality is large and the environment is non-stationary \cite{woergoetter_rl_2008}. Also, DL's capability to learn complex patterns is sometimes prone to misclassification \cite{nguyen_dl_fool_2015}.
To mitigate, in recent years, RL algorithms have been successfully combined with deep NN \cite{li_deeprl_review_2017} giving rise to novel learning strategies. This integration has been used either in approximating RL functions using deep NN architectures or in training deep NN using RL. 

The first notable example of such an integration is the Deep Q-network (DQN) \cite{mnih_human-level_2015} which combines Q-learning with deep NN. The DQN agent, when presented with high-dimensional inputs, can successfully learn policies using RL. The action-value function is approximated for optimality using deep CNN. The deep CNN, using experience replay and target network, overcomes the instability and divergence sometimes experienced while approximating Q-function with shallow NN.

Another deep RL algorithm is the Double DQN which is an extension of the DQN algorithm \cite{van_hasselt_deep_2015}. In certain situations the DQN suffers from substantial overestimations inherited from the implemented Q-learning which are overcome by replacing the Q-learning of the DQN with a double Q-learning algorithm \cite{NIPS2010_3964_Hasselt10}. The DQL learns two value functions, by assigning an experience randomly to update one of them, resulting in two sets of weights. During every update one set determines the greedy policy while the other its value.
Other deep RL algorithms include: Deep Deterministic Policy Gradient, Continuous DQN, Asynchronous N-step Q-learning, Dueling network DQN, Prioritized Experience Replay, Deep SARSA, Asynchronous Advantage Actor-Critic, and Actor-Critic with Experience Replay \cite{li_deeprl_review_2017}.

\section{Applications to Biological Data}
\label{sec-applications}

The techniques outlined above, also available as open-source tools (e.g., see \cite{erickson_toolkits_2017} for a mini review on tools based on DL), have been used in mining Biological data. The applications, as reported in the literature, are provided below for data coming from each of the application domains.

Table \ref{tab:main} summarizes the state-of-the art applications of DL and RL to biological data (see Fig. \ref{fig_1} H). It also reports on individual applications in each of these domains and the data type on which the methods have been applied.

\begin{sidewaystable}[!htbp]
  \centering
  \caption{Summary of [Deep] [Reinforcement] Learning Applications to Biological Data}
    \begin{tabular}
    {cllcccccc}
\hline
\multicolumn{1}{c}{\multirow{3}[0]{*}{{\begin{tabular}[c]{@{}c@{}}App.\\Dom.\end{tabular}}}} & \multicolumn{1}{c}{\multirow{3}[0]{*}{Purpose}} & \multicolumn{1}{c}{\multirow{3}[0]{*}{{\begin{tabular}[c]{@{}c@{}}Data\\Type\end{tabular}}}} & \multicolumn{6}{c}{Method \& Reference} \\
\cline{4-9}
\multicolumn{1}{c}{} &       &       & \multicolumn{5}{c}{DL}                       & \multicolumn{1}{c}{\multirow{2}[0]{*}{RL}} \\
\cline{4-8}
\multicolumn{1}{c}{} &       &       & \multicolumn{1}{c}{DA} & \multicolumn{1}{c}{DBN} & \multicolumn{1}{c}{DNN} & \multicolumn{1}{c}{CNN} & \multicolumn{1}{c}{RNN$^\psi$} &       \\
    \hline
    \multicolumn{1}{c}{\multirow{5}[0]{*}{\begin{sideways}Omics\end{sideways}}} & \multicolumn{1}{c}{\multirow{5}[0]{*}{\begin{tabular}[l]{@{}l@{}}GE, DD,\\DmsPS,\\PrSD,\\DrD,\\PsDs\end{tabular}}} & \multirow{5}[2]{*}{\begin{tabular}[c]{@{}l@{}}SJP/Se,\\miRNA,\\Prtn,\\Gene\end{tabular}} & \multicolumn{1}{c}{\multirow{5}[0]{*}{\cite{fakoor2013,danaee2016,li_2016}}} & \multicolumn{1}{c}{\multirow{5}[0]{*}{\cite{Lee3045382,6944490,Chen2015,pan_rbm_2017}}} & \multicolumn{1}{c}{\multirow{5}[0]{*}{\cite{DBLPChenLNSX16,doigkv1025,Angermueller055715,doibtu703,heffernan2015,Tian201664,wan_cpi_dnn_2016,yuan_deepgene_2016,hamanaka_cgbvs_cpi_dnn_2017}}} & \multicolumn{1}{c}{\multirow{5}[0]{*}{\cite{Denas2013DeepMO,alipanahi_deepbind_2015,Kelley2016,DBLPZengELG16,citeulike13721890,Huang069682,DBLPWang0MX15}}} & \multicolumn{1}{c}{\multirow{5}[0]{*}{\cite{DBLPParkMCY16,DBLPLeeBPY16}}} & \multirow{5}[2]{*}{\cite{5695478,5715222,bocicor_rl_dna_2011,6999302}} \\
          &       &       &       &       &       &       &       &              \\
          &       &       &       &       &       &       &       &              \\
          &       &       &       &       &       &       &       &              \\
          &       &       &       &       &       &       &       &              \\
    \hline
    \multicolumn{1}{c}{\multirow{5}[2]{*}{\begin{sideways}Bioimaging\end{sideways}}} & \multicolumn{1}{c}{\multirow{5}[0]{*}{\begin{tabular}[c]{@{}l@{}}Seg,\\BCaD,\\ CACI,\\CCnt\end{tabular}}} & \multirow{5}[0]{*}{EMI} & \multirow{5}[2]{*}{\cite{DBLP0005XLGWTM16}} & \multirow{5}[2]{*}{} & \multicolumn{1}{c}{\multirow{5}[2]{*}{\cite{xu_deep_2014,chen2016}}} & \multicolumn{1}{c}{\multirow{5}[0]{*}{\cite{1495508,Ciresan2013,NIPS2012_4741,Prnamaa050757,Ferrari2017629,DBLPKrausBF15,Eulenberg081364,7405812}}} & \multirow{5}[0]{*}{} & \multirow{5}[0]{*}{} \\
          &       &       &       &       &       &       &       &              \\
          &       &       &       &       &       &       &       &              \\
          &       &       &       &       &       &       &       &              \\
          &       &       &       &       &       &       &       &              \\
    \hline
    \multicolumn{1}{c}{\multirow{4}[0]{*}{\begin{sideways}{\begin{tabular}[c]{@{}c@{}}Medical\\Imaging\end{tabular}}\end{sideways}}} & \multicolumn{1}{c}{\multirow{4}[0]{*}{\begin{tabular}[c]{@{}l@{}}Seg,\\ID,\\DD\end{tabular}}} & [f/s]MRI & \multicolumn{1}{c}{\cite{7420737,suk_da_2013,shi_df_AD_2017}} & \multicolumn{1}{c}{\cite{broschtam2013,Suk2014569,Li2014,10338900229}} & \multicolumn{1}{c}{\cite{DBLPHavaeiDWBCBPJL15}\cite{shi_sdpn_ad_2017}$^\upsilon$} & \multicolumn{1}{c}{\cite{DBLPHavaeiGLJ16,DBLPHosseiniAslGE16,Kleesiek2016460,Sarraf070441,Nie2016,kamnitsas_3dcnn_2017}} & \multicolumn{1}{c}{\cite{DBLPStollengaBLS15}} &       \\
\cline{3-9}
&       & CT   & \multicolumn{1}{c}{\cite{NIPS2013_5030}} &       & \multicolumn{1}{c}{\cite{Lerouge20152847}} & \multicolumn{1}{c}{\cite{Fritscher2016,Gao201749,DBLPChoLSCD15,DBLPRothLLYSKKS15,chengchen2016,DBLPShinRGLXNYMS16,shen_cnn_ln_2017,tajbakhsh_cnn_ln_2017}} &       &       \\
\cline{3-9}
&       & PET   & \multicolumn{1}{c}{\cite{suk_da_2013}} & \multicolumn{1}{c}{\cite{Suk2014569,Li2014,7062868}} &  \multicolumn{1}{c}{\cite{shi_sdpn_ad_2017}$^\upsilon$}     & \multicolumn{1}{c}{\cite{Ypsilantis7036}} &       &       \\
\cline{3-9}
&       & OthI & \multicolumn{1}{c}{\cite{DBLPGondara16}} & \multicolumn{1}{c}{\cite{ngo_dl_hrt_2017}}      & \multicolumn{1}{c}{\cite{7090943}} & \multicolumn{1}{c}{\cite{chengchen2016,7422082,7401052,7419037,7348440,arevalo_cnn_brst_2016,jiao_cnn_brst_2016,sun_cnn_brst_2017,kooi_cnn_brst_2017,dhungel_dl_brst_2017,sirinukunwattana_gl_2017,dou_heart_2017}} & \multicolumn{1}{c}{\cite{DBLPStollengaBLS15}} & \cite{Sahba2008} \\
    \hline
    \multicolumn{1}{c}{\multirow{4}[0]{*}{\begin{sideways}BMI\end{sideways}}} & \multicolumn{1}{c}{\multirow{4}[0]{*}{\begin{tabular}[c]{@{}l@{}}MA,\\MD,\\ER,\\CogS\end{tabular}}} & EEG   & \multicolumn{1}{c}{\cite{li_da_mi_2014,jirayucharoensak2014}} & \multicolumn{1}{c}{\cite{lu_rbm_mi_2016,An2014,li_dbn_as_2013,7033556,zheng2015investigating,wulsin2011,Zhao2015,langkvist2012}\cite{chai_dbn_driver_2017}$^\gamma$} & \multicolumn{1}{c}{\cite{DBLPSturmBSM16,kumar_dnn_mid_2016}} & \multicolumn{1}{c}{\cite{yang2015,tabar_cnn_mi_eeg_2017,sakhavi_mi_2015,IAAI1715007,7383844,Mirowski20091927}} & \multicolumn{1}{c}{\cite{6890301,DBLPBashivanRYC15,Petrosian2000201,journalsDavidsonJP07}} & \cite{Lampe2557533,bauer2015} \\
\cline{3-9}
&       & EMG   &       &       &       & \multicolumn{1}{c}{\cite{DBLPAtzoriCM16,7457459}} &       &      \\
\cline{3-9}
&       & ECG   &       & \multicolumn{1}{c}{\cite{7023547,DBLPYanQWZ0W15,wu_ecg_dbn_2016}} & \multicolumn{1}{c}{\cite{rahhal_dnn_ecg_2016}} &       &       &       \\
\cline{3-9}
&       & NS &       &       &       &       &       & 
\cite{digiovanna_coadaptive_RL_2009,mahmoudi_symbiotic_2011,sanchez_RL_test_2011,mahmoudi_RL_2013,wangzheng2013,pohlmeyer2014,wang_QRL_2017} \\
    \hline
    \end{tabular}%
  \label{tab:main}%
 
 \vspace{2ex}
 
 \raggedright Legends: App. Dom.--Application Domain; $^\psi$--including LSTM; $^\upsilon$--using DPN; $^\gamma$--using Sparse-DBN; \underline{\textbf{Abbreviations in `App' column}}: GE--Gene Expression; DD--Disease Diagnosis (including Alzheimer's/ Huntington/ Schizophrenia/ Mental/ Emotional State); DmsPS--DNA methylation state prediction \cite{Angermueller055715}, Pathogenicity \cite{doibtu703}, Sequence assembly \cite{5695478,5715222,bocicor_rl_dna_2011}; PrSD--Protein Structure (binding) Determination; DrD--Drug Discovery; Seg--Segmentation; BCaD--Breast Cancer Detection; CACI--Classification \& Analysis of Cell Image; CCnt--Cell Count; ID--Image Denoising;  MA--Motor Action decoding; MD--Movement Decoding; ER--Emotion Recognition; CogS--Cognitive State Determination;\\
\underline{\textbf{Abbreviations in `Data Type' column}}: SJP--Splice Junction Prediction; Se--[DNA/RNA/ChIP/DNase] sequence \& microarray GE; Prtn--Protein properties; OthI--Images not used elsewhere (e.g., UlS/EMI/EnI/XRay/CFI/MMM/cardiac MRI); NS--Neural Spikes.
\end{sidewaystable}%

\subsection{Omics}
\label{subsec-app-omics}
Some DL and RL methods have been extensively used in \textit{Omics} (such as genomics, proteomics or metabolomics) research to extract features, functions, structure, and molecular dynamics from the raw biological sequence data (e.g., DNA, RNA, and amino-acids). 
Specifically, mining sequence data is a challenging task. Different analyses (e.g., gene expression profiling, splicing junction prediction, sequence specificity prediction, transcription factor determination, protein-protein interaction evaluation, etc.) dealing with different types of sequence data have been reported in the literature.

To identify splicing junction at DNA level, a tedious job to do manually, Lee et al. proposed a DBN based unsupervised method to perform the auto-prediction \cite{Lee3045382}. 
Profiling gene expression (GE) is a demanding job. Chen et al. exploited a DNN based method for GE profiling on RNA-seq and microarray-based GE Omnibus dataset \cite{DBLPChenLNSX16}. 
The ChIP-seq data were preprocessed, using CNN, into a 2D matrix where each row denoted a gene's transcription factor activity profile \cite{Denas2013DeepMO}. 
Also, somatic point mutation based cancer classification was performed using DNN \cite{yuan_deepgene_2016}.
In addition, DA based methods have been used for feature extraction in cancer diagnosis and classification (Fakoor et al. used sparse DA method \cite{fakoor2013}) in combination with related gene identification (Danaee et al. used stacked Denoising DA \cite{danaee2016}) from GE data.

Alipanahi et al. used Deep CNN structure to predict DNA- and RNA-binding proteins' ([D/R]BPs) role in alternative splicing and examined the effect of disease associated genetic variants (GV) on transcription factor binding and GE \cite{alipanahi_deepbind_2015}. 
Zhang et al. developed a DNN framework to model structural features of RBPs \cite{doigkv1025}. Pan et al. proposed a hybrid CNN-DBN model to predict RBP interaction sites and motifs on RNAs \cite{pan_rbm_2017}.
Quang et al. proposed a DNN model to annotate and identify pathogenicity in GV  \cite{doibtu703}. 

Identifying the best discriminative genes/microRNAs (miRNAs) is a challenging task. Ibrahim et al. proposed a group feature selection method from genes/miRNAs based on expression profile using DBN and active learning \cite{6944490}. CNN was used to 
interpret noncoding genome by annotating them \cite{Kelley2016}. Also, Zeng et al. employed CNN to predict the binding between DNA and protein \cite{DBLPZengELG16}. Zhou et al. proposed a CNN based approach to identify noncoding GV \cite{citeulike13721890} which was also used by Huang et al. for a similar purpose \cite{Huang069682}.
Park et al. proposed a LSTM based tool to automatically predict miRNA precursor \cite{DBLPParkMCY16}. Also, Lee et al. presented a deep RNN framework for automatic miRNA target prediction \cite{DBLPLeeBPY16}.

DNA methylation (DM) causes DNA segment activity alteration without affecting the sequence, thus, detecting it's state in a sequence is important. Angermueller et al. used DNN based method to estimate DM state by predicting the changes in single nucleotides and uncovering sequence motifs \cite{Angermueller055715}. 

Proteomics pose many complex computational problems to solve. Estimating complete protein structures from biological sequences, in 3D space, is a complex and NP hard problem. Alternatively, the protein structures can be divided into independent sub-problems (e.g., torsion angle, access surface area, dihedral angles, etc.) and solved in parallel, and estimate the secondary protein structures (2-PS). Predicting compounds-protein interaction (CPI) is very interesting from drug discovery point of view and tough to solve. 

Heffernan et al. proposed an iterative DNN scheme to solve these sub-problems 
for 2-PS \cite{heffernan2015}. Wang et al. utilized deep CNN to predict 2-PS \cite{DBLPWang0MX15}. Li et al. proposed DA learning based model to reconstruct protein structure based on a template \cite{li_2016}.
Also, DNN based methods to predict CPI \cite{Tian201664,wan_cpi_dnn_2016,hamanaka_cgbvs_cpi_dnn_2017} have also been reported. 

In medicine, model organisms are often used for translational research. Chen et al. used bimodal DBNs to predict responses of human cells under certain stimuli based on responses of rat cells obtained with same stimuli \cite{Chen2015}.

RL has also been used in omics, for example, Yang et al. used binary particle swarm optimization and RL to predict bacterial genomes \cite{5695478}, Ralha et al. used RL through a system called BioAgent to increase the accuracy of biological sequence annotation \cite{5715222}, and Bocicor et al. solved the problem of DNA fragment assembly using RL based framework \cite{bocicor_rl_dna_2011}.
Zhu et al. proposed hybrid RL method, with text mining, for constructing protein-protein interaction networks \cite{6999302}.

\subsection{Bioimaging}
In biology, DL architectures targeted on pixel levels of a biological image to train the NN. Ning et al. used CNN for pixel-wise image segmentation of nucleus, cytoplasm, cell, and nuclear membranes using Electron Microscope Image (EMI) \cite{1495508}. Reduced pixel noise and better abstract features of biological images can be obtained by adding multiple layers. Ciresan et al. employed deep convolutional neural networks to identify mitosis in histology images of the breast \cite{Ciresan2013}, and similar architecture was also used to find neuronal membranes and automatically segment neuronal structures in EMI \cite{NIPS2012_4741}. Xu et al. used Stacked Sparse DA architecture to identify nuclei in the histopathology images of the breast cancer \cite{DBLP0005XLGWTM16}. Xu et al. classified Colon cancer images using Multiple Instance Learning (MIL) from DNN learnt features \cite{xu_deep_2014}.

Besides pixel level analysis, DL have also been applied to cell and tissue level analysis. 
Chen et al. employed DNN in label-free cell classification \cite{chen2016}. P{\"a}rnamaa and Leopold used CNN  to automatically detect fluorescent protein in various subcellular localization patterns using microscopy images of yeast \cite{Prnamaa050757}.  Ferrari et al. used CNNs to count bacterial colonies in agar plates \cite{Ferrari2017629}. Kraus et al. integrated both the segmentation as well as classification in a model which can be utilized to classify the microscopy images of the yeast \cite{DBLPKrausBF15}. Flow cytometry is used in cellular biology through cycle analysis to monitor different stages of a cell-cycle. Eulenberg et al. proposed deep flow model, combining non-linear dimension reduction with CNN, to analyze single cell flow cytometry images \cite{Eulenberg081364}. Furthermore, CNN architecture was employed to segment and recognize neural stem cells in images taken by bright field microscope \cite{7405812}, and DBN for analyzing Gold immunochromatographic strip \cite{zeng_deep_2016}.

\subsection{Medical Imaging}
DL and RL architectures have been widely used in analyzing medical images obtained from--  magnetic resonance ([f/s]MRI), CT scan, positron emission tomography (PET), radiography/ fundus (e.g., X-ray, CFI), microscope, ultrasound (UlS)-- to denoise, segment, classify, detect anomalies and diseases from these images. 

Segmentation is a process of partitioning an image based on some specific patterns. 
Sirinukunwattana et al. reported the results of the Gland Segmentation competition from colon histology images \cite{sirinukunwattana_gl_2017}. Kamnitsas et al. proposed 3D dual pathway CNN to simultaneously process multi-channel MRI and segment lesions related to  tumors, traumatic injuries, and ischemic stroke \cite{kamnitsas_3dcnn_2017}.
Stollenga et al. segmented neuronal structures from 3D EMI and brain MRI using multi dimensional RNN \cite{DBLPStollengaBLS15}. Fritscher et al. used deep CNN for volume segmentation from head-neck region's CT scans  \cite{Fritscher2016}. Havaei et al. segmented brain tumor from MRI using CNN \cite{DBLPHavaeiGLJ16}, and  DNN \cite{DBLPHavaeiDWBCBPJL15}. 
Brosch and Tam proposed a DBN based manifold learning method of 3D brain MRI \cite{broschtam2013}. Cardiac MRIs were segmented for heart's left ventricle using DBN \cite{ngo_dl_hrt_2017}, and blood pool and myocardium using CNN \cite{dou_heart_2017}. 
Mansoor et al. automatically segmented anterior visual pathway from MRI sequences using stacked DA model \cite{7420737}. Lerouge et al. proposed DNN based method to label CT scans \cite{Lerouge20152847}. 

Success of many medical image analysis methods depends on image denoising. Gondara proposed a denoising technique utilizing  convolutional denoising DA, and validated it with mammograms and dental radiography \cite{DBLPGondara16}. While Agostinelli et al.
presented an adaptive multi-column stacked sparse denoising autoencoder (DA) method for image denoising which was validated using CT Scan images of the head \cite{NIPS2013_5030}. 

Detecting anomaly in medical images is widely used for disease diagnosis. 
Several models were applied to detect Alzheimer's Disease (AD) and Mild Cognitive Impairment (MCI) from MRI and PET scans including DA \cite{suk_da_2013,shi_df_AD_2017}, DBM \cite{Suk2014569}, RBM \cite{Li2014}, and multimodal stacked deep polynomial network (MM-SDPN) \cite{shi_sdpn_ad_2017}.

Due to its facilitating structure, CNN has been the most popular DL architecture for image analysis. CNN was applied to classify breast masses from mammograms (MMM) \cite{arevalo_cnn_brst_2016,jiao_cnn_brst_2016,sun_cnn_brst_2017,kooi_cnn_brst_2017,dhungel_dl_brst_2017}, diagnose AD using different neuroimages (e.g., brain MRI \cite{DBLPHosseiniAslGE16}, brain CT scans \cite{Gao201749}, and (f)MRIs \cite{Sarraf070441}), and rheumatoid arthritis from hand radiographs \cite{7348440}. 
CNN was also used extensively: on CT scans to detect-- anatomical structure \cite{DBLPChoLSCD15}, sclerotic metastases of spine along with colonic polyps and lymph nodes (LN) \cite{DBLPRothLLYSKKS15}, thoracoabdominal LN and interstitial lung disease (ILD) \cite{DBLPShinRGLXNYMS16}, pulmonary nodules \cite{chengchen2016,shen_cnn_ln_2017,tajbakhsh_cnn_ln_2017}; 
on (f)MRI and diffusion tensor images to extract deep features for brain tumor patients' survival time prediction \cite{Nie2016}; on MRI to detect neuroendocrine carcinoma \cite{Kleesiek2016460}; on UlS images to diagnose Breast Lesions \cite{chengchen2016} and ILD \cite{7422082}; on CFI to detect hemorrhages \cite{7401052}; on endoscopy images to diagnose digestive organ related diseases \cite{7419037}; on PET images to identify oesophagal carcinoma and predict responses of neoadjuvant chemotherapy \cite{Ypsilantis7036}.

In addition, DBN was successfully applied to identify: Attention Deficit Hyperactivity Disorder \cite{7062868}, and Schizophrenia (SZ) and Huntington Disease from (f/s)MRI \cite{10338900229}. And, a DNN based method was proposed to successfully identify the fetal abdominal standard plane in UlS images \cite{7090943}.

RL was used in segmenting transrectal UlS images to estimate location and volume of the prostate \cite{Sahba2008}.

\subsection{[Brain/Body]-Machine Interfaces}
\label{subsec-app-bmi}
DL and RL methods have been applied to BMI signals (e.g., electroencephalogram, EEG; electrocardiogram, ECG; electromyogram, EMG) mainly from (brain) function decoding and anomaly detection perspectives.

Various DL architectures have been used in classifying EEG signals to decode Motor Imagery (MoI). CNN was applied in the classification pipeline using -- augmented common spatial pattern features which covered various frequency ranges \cite{yang2015}; features based on combined selective location, time, and frequency attributes which were then classified using DA \cite{tabar_cnn_mi_eeg_2017}; and signal's dynamic energy representation \cite{sakhavi_mi_2015}.
DBN was also employed-- in combination with softmax regression to classify signal frequency information as features \cite{lu_rbm_mi_2016}; and in conjunction with Ada-boost algorithm to classify single channels \cite{An2014}.
DNN was used-- with variance based common spatial pattern (CSP) features to classify MoI EEG \cite{kumar_dnn_mid_2016}, and to find neural patterns occurring at each time points in single trials where the input heatmaps were created with layer-wise relevance propagation technique \cite{DBLPSturmBSM16}.
In addition, MoI EEG signals were classified by denoising DA using multifractal attribute features \cite{li_da_mi_2014}.

DBN was used by Li et al. to extract low dimensional latent features as well as critical channel selection which led to an early framework for affective state classification using EEG signals \cite{li_dbn_as_2013}. In a similar work, Jia et al. used semi-supervised approach with an active learning to train DBN and generative RBMs for the classification \cite{7033556}. Later, using differential entropy as features to train DBN, Zheng et al. examined dominant frequency bands and channels of EEG in an emotion recognition system \cite{zheng2015investigating}. 
Jirayucharoensak et al. used PCA extracted power spectral densities from each EEG channel, which were corrected by covariate shift adaptation to reduce non-stationarity, as features to stacked DA to detect emotion \cite{jirayucharoensak2014}. 
Tripathi et al. explored DNN (with Softmax activator and Dropout) and CNN \cite{li_hierarchical_2017}  (with Tan Hyperbolic, Max Pooling, Dropout, and Softplus) for emotion classification from the DEAP dataset using EEG signals and response face video \cite{IAAI1715007}. 
Using similar data from the MAHNOB-HCI dataset, Soleymani et al. detected continuous emotion using RNN-LSTM \cite{6890301}. 
Channel-wise CNN \& its variant with RBM \cite{7383844}, and AR-model based features with sparse-DBN \cite{chai_dbn_driver_2017}, was used to estimate driver's cognitive states using EEG data. 

In another approach to model cognitive events, EEG signals were transformed to time-lagged multi-spectral images and fed to CNN for learning the spectral and spatial representations of each image, followed by an adapted RNN (LSTM) to find the temporal patterns in the image sequence \cite{DBLPBashivanRYC15}.

DBN has been employed in classifying EEG signals for anomaly detection in diverse scenarios including: online waveform classification \cite{wulsin2011}; AD diagnosis \cite{Zhao2015}; integrated with HMM to understand sleep phases \cite{langkvist2012}. To detect and predict seizures-- CNN was used through classification of synchronization patterns \cite{Mirowski20091927}; RNN predicted specific signal features related to seizure after being trained with data preprocessed by wavelet decomposition \cite{Petrosian2000201}. 
Also, a lapse of responsiveness warning system was proposed using RNN (LSTM) \cite{journalsDavidsonJP07}.

Using CNN Park \& Lee \cite{7457459} and Atzori et al. \cite{DBLPAtzoriCM16} decoded hand movements from EMG signals. 

ECG Arrhythmias were successfully detected using DBN \cite{wu_ecg_dbn_2016} and DNN \cite{rahhal_dnn_ecg_2016}. DBN was also used to classify ECG signals acquired with two-leads 
\cite{DBLPYanQWZ0W15}, and in combination with nonlinear SVM and Gaussian kernel \cite{7023547}.


RL has also been applied in BMI research. Concentrating mainly on controlling (prosthetic/robotic) devices, several studies have been reported, including: mapping neural activity to intended behavior through coadaptive BMI (using TD($\lambda$)) \cite{digiovanna_coadaptive_RL_2009} and symbiotic BMI (using actor-critic) \cite{mahmoudi_symbiotic_2011}, a testbed targeting center-out reaching task in primates for creating more realistic BMI control models \cite{sanchez_RL_test_2011}, Hebbian RL for adaptive control by mapping neural states to prosthetic actions \cite{mahmoudi_RL_2013}, BMI for unsupervised decoding of cortical spikes in multistep goal-directed tracking task (using Q($\lambda$)) \cite{wangzheng2013}, adaptive BMI capable of adjusting to dramatic reorganizing neural activities with minimal training and stable performance over long duration (using actor-critic) \cite{pohlmeyer2014}, BMI for efficient nonlinear mapping of neural states to actions through sparsification of state-action mapping space using quantized attention-gated kernel RL as an approximator \cite {wang_QRL_2017}. Also, Lampe et al. proposed BMI capable of transmitting imaginary movements evoked EEG signals over the Internet to remotely control robotic device \cite{Lampe2557533}, and Bauer and Gharabaghi combined RL with Bayesian model to select dynamic thresholds for improved performance of restorative BMI \cite{bauer2015}.

%
\begin{figure}[!tbh]
	\hspace{-3cm}  
    \includegraphics{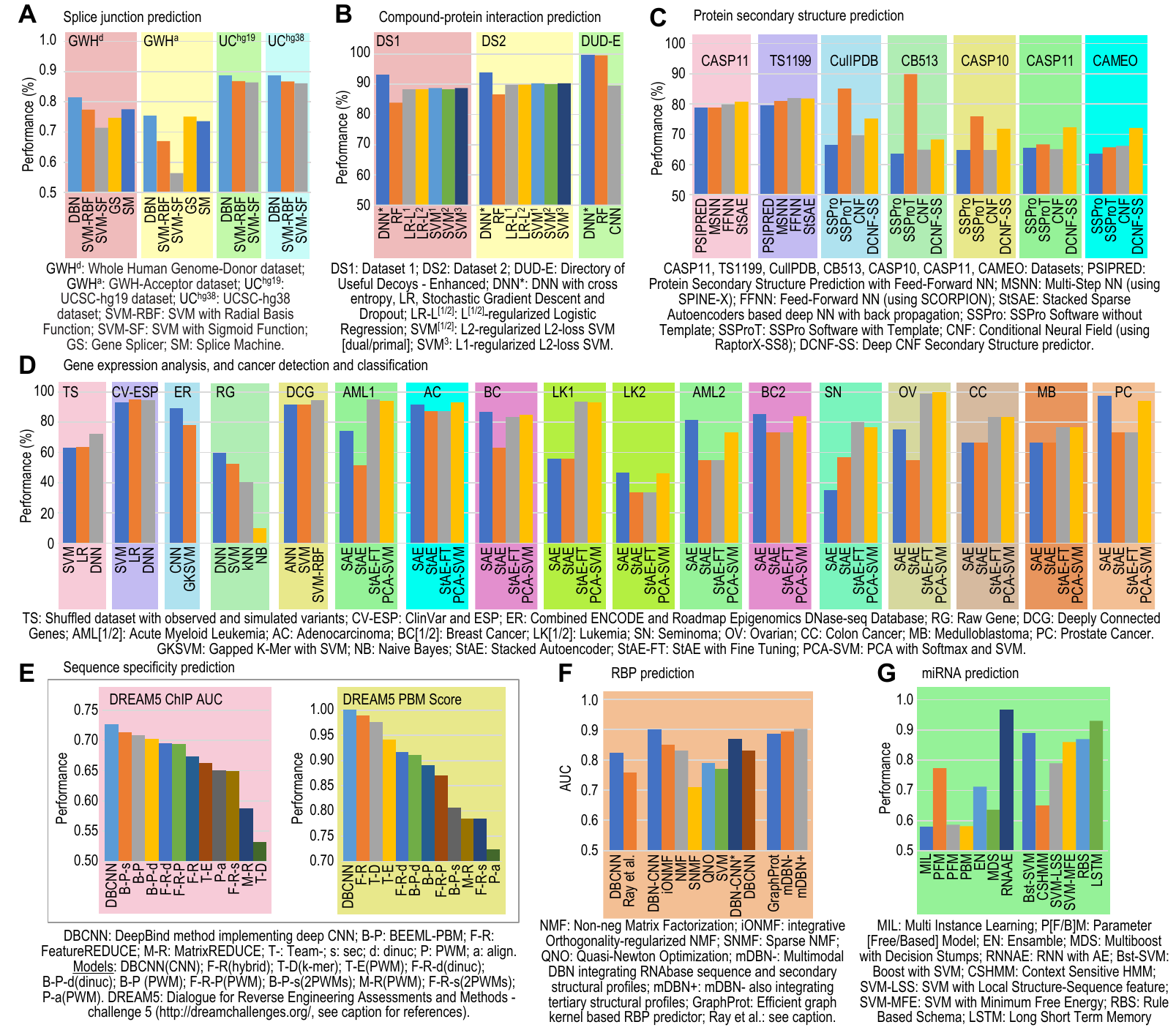}
  \caption{
Performance comparison of representative DL techniques when applied to Omics data in: predicting splice junction (\textbf{A}), compound-protein interactions (\textbf{B}), and secondary/tertiary structures of proteins (\textbf{C}); analyzing gene expression data, and classifying and detecting cancers from them (\textbf{D}); and predicting DNA- and RNA-sequence specificity (\textbf{E}), RNA binding proteins (\textbf{F}), and micro RNA precursors (\textbf{G}). Ray et al.: method proposed in \cite{ray_rbp_2013}.}
\label{fig_omics}

\end{figure}

\section{Performance Analysis and Comparison}
\label{sec-perf-comp}
Comparative test results, in the form of performances/ accuracies of each DL technique when applied to the data coming from \textit{Omics} (Fig. \ref{fig_omics}), \textit{Bioimaging} (Fig. \ref{fig_bioi}), \textit{Medical Imaging} (Fig. \ref{fig_medi}), and \textit{[Brain/Body]-Machine Interfaces} (Fig \ref{fig_bmi}), are summarized below to facilitate the reader in selecting the appropriate method for h[is/er] research. The reported performances can be regarded as a metric to evaluate the strengths/ weaknesses of a particular technique with a given set of parameters on a specific dataset.
It should be noted that several factors (e.g., data pre-processing, network architecture, feature selection and learning,  parameters' optimization, etc.) collectively determine the accuracy of a method.

In Figs. \ref{fig_omics}--\ref{fig_bmi}, each group of bars indicates accuracies/performances of comparable DL or non-DL techniques when applied to same data and reported in an individual study. And, each bar in a group shows the (mean) performance of different runs of a technique on either multiple subjects/datasets (for means, error bar is $\pm$ standard deviation).

\subsection{Omics}
\label{subsec-omics-perf}
Fig. \ref{fig_omics}\textbf{A} reports that DBN outperforms other methods in predicting splice junction when applied to: two datasets from Whole Human Genome database (GWH-donor, GWH-acceptor) and two from UCSC genome database (UCSC-hg19, UCSC-hg38) \cite{Lee3045382}. In the GWH datasets, DBN based method achieved superior F1-score  (0.81 and 0.75) against SVM based methods with Radial Basis (0.77 and 0.67) and Sigmoid (0.71 and 0.56) Functions, and other splicing techniques like Gene Splicer (0.74 and 0.75) and Splice Machine (0.77 and 0.73). Also, in the UCSC datasets, DBN achieved highest classification accuracy (0.88 and 0.88) in comparison to SVM-RBF (0.868 and 0.867) and SVM-SF (0.864 and 0.861).

Performance comparison of CPI is shown in Fig. \ref{fig_omics}\textbf{B}. Tested over two CPI datasets, a DNN based method (DNN*) achieved superior prediction accuracy (93.2\% in dataset1 and 93.8\% in dataset2) compared to other methods based on RF (83.9\% and 86.6\%), LR (88.3\% and 89.9\% using LR$^2$), and SVM (88.7\% and 90.3\% using SVM$^3$) \cite{Tian201664}. In another study, a similar DNN* was applied on DUD-E dataset where it achieved higher accuracy (99.6\%) over RF (99.58\%) and CNN (89.5\%) based methods \cite{wan_cpi_dnn_2016}. As per the accuracies reported in \cite{Tian201664}, the RF based method had lower values in comparison to the LR and SVM based methods which had similar values. Whereas, when applied on DUD-E dataset (reported in \cite{wan_cpi_dnn_2016}), the RF based method outperforms the CNN based method. This may be attributed to the fact that, classification problems are data dependent and despite being one of the best classifiers \cite{fernandez_rf_best_2014}, RF performs poorly on the DUD-E dataset.

In predicting 2-PS, DL based methods outperforms other methods (see Fig. \ref{fig_omics}\textbf{C}). When applied on two datasets (CASP11 and TS1199), the stacked sparse autoencoder (StSAE) based method achieved superior prediction accuracy (80.8\% and 81.8\%) in comparison to other NN based methods (FFNN: 79.9\% and 82\%, MSNN: 78.8\% and 81\%, and PSIPRED: 78.8\% and 79.7\%) \cite{heffernan2015}. Another DL method with Conditional Neural Fields (DLCNF), when tested on five different datasets (CullPDB53, CB513, CASP1054, CASP1155, CAMEO), better predicted the 2-PS (Q8 accuracy: 75.2\%, 68.3\%, 71.8\%, 72.3\%, 72.1\%) in comparison to other non-template based methods (SSPro: 66.6\%, 63.5\%, 64.9\%, 65.6\%, 63.5\%; CNF: 69.7\%, 64.9\%, 64.8\%, 65.1\%, 66.2\%) \cite{DBLPWang0MX15}. However, when a template of solved protein structure from PDB was used, the SSpro with template obtained the best accuracy (SSProT: 85.1\%, 89.9\%, 75.9\%, 66.7\%, 65.7\%).

To annotate GV in identifying pathogenic variants from two datasets (TS and CVESP in Fig. \ref{fig_omics}\textbf{D}), a DNN based method performed better (72.2\% and 94.6\%) than LR (63.5\% and 95.2\%) and SVM (63.1\% and 93.0\%) based methods. Another CNN based approach to predict DNA sequence accessibility was tested on data from ENCODE \& REC databases and was reported to outperform gapped k-mer SVM method (mean AUC of 0.89 vs. 0.78) \cite{Kelley2016}. In classifying cancer based on somatic point mutation using raw TCGA data containing 12 cancer types, a DNN based method outperformed non-DL methods (60.1\% vs. [SVM: 52.7\%, kNN: 40.4\%, NB: 9.8\%]) \cite{yuan_deepgene_2016}. To detect breast cancer using GE data from TCGA database, a Stacked Denoising Autoencoder (StDAE) was employed to extract features. According to the reported accuracies of non-DL classifiers (ANN, SVM, and SVM-RBF), StDAE outperformed other feature extraction methods such as PCA and KPCA (SVM-RBF classification accuracies for StDAE, PCA, and KPCA were 98.26\%, 89.13\%, and 97.32\%, respectively) \cite{danaee2016}. Also, deeply connected genes were better classified with StDAE extracted features (accuracies- ANN: 91.74\%, SVM: 91.74\%, and SVM-RBF: 94.78\%) \cite{danaee2016}. Another study on classifying cancer, using 13 different GE datasets taken from the literature, reported that the use of PCA in data dimensionality reduction, before applying SAE, StAE, and StAE-FT for feature extraction, facilitates more accurate extraction of features (except AC and OV in Fig. \ref{fig_omics}\textbf{D}) for classification using SVM with Gaussian kernel \cite{fakoor2013}.

Sequence specificities of [D/R]BPs prediction was performed more accurately using a deep CNN based method in comparison to other non-DL methods participated at the DREAM5\footnote{DREAM5 challenge: \url{http://dreamchallenges.org/} and \cite{marbach_dream5_2012,weirauch_dream5_2013}.} challenge \cite{alipanahi_deepbind_2015}. As seen in Fig. \ref{fig_omics}\textbf{E}, the CNN based method (DeepBind) outperformed other methods in ChIP AUC values (top two values- DeepBind: 0.726 vs. BEEML-PBM\_sec: 0.714) and PBM scores (two top scores- DeepBind: 0.998 vs. FeatureREDUCE: 0.985) \cite{marbach_dream5_2012,weirauch_dream5_2013}.

Moreover, in predicting RBP, DL based methods outperformed non-DL methods as seen in Fig. \ref{fig_omics}\textbf{F}. As reported in using CLIP AUC values, DBCNN outperforms Ray et al. (0.825 vs. 0.759) \cite{alipanahi_deepbind_2015}, multimodal DBN outperforms GraphProt (0.902 vs. 0.887) \cite{doigkv1025}, and DBN-CNN hybrid outperforms NMF based methods (0.9 vs. 0.85) \cite{pan_rbm_2017}.

Also, in predicting miRNA precursor (Fig. \ref{fig_omics}\textbf{G}), RNN with DA outperformed other non-DL methods (RNNAE: 0.97 vs. [MIL: 0.58, PFM: 0.77, PFM: 0.59; PBM: 0.58, EN: 0.71, and MDS: 0.64]) \cite{DBLPLeeBPY16}. And LSTM outperformed SVM methods (LSTM: 0.93 vs. [Boost-SVM: 0.89, CSHMM: 0.65, SVM-LSS: 0.79, SVM-MFE: 0.86, and RBS: 0.8]) \cite{DBLPParkMCY16}.

\subsection{Bioimaging}
\label{subsec-bioi-perf}
\begin{figure}[!htb]
\hfill\includegraphics{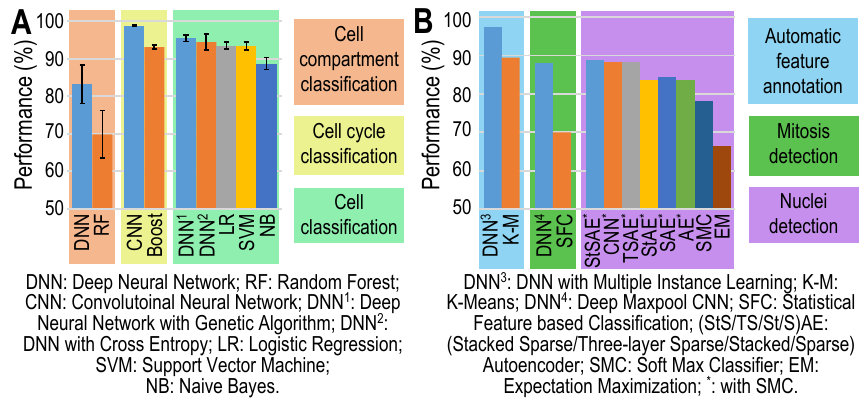}
\caption{
 Performance comparison of some DL and conventional ML techniques when applied to Bioimaging application domain. 
\textbf{A}. Performances in classifying electron microscope images for cell compartments, cell cycles, and cells. \textbf{B}. Performances in analyzing images to automatically annotate features, and detect mitosis and cell nuclei.}

\label{fig_bioi}
\end{figure}

DNN was used in detecting 12 different cellular compartments from microscopy images and was reported to have achieved classification accuracy of 87\% compared to 75\% for RF \cite{Prnamaa050757}. The mean performance of the detection was 83.24$\pm$5.18\% using DNN and 69.85$\pm$6.33\% using RF (Fig. \ref{fig_bioi}\textbf{A}).
 In classifying flow cytometry images for cell cycle phases, Deep CNN with non-linear dimension reduction outperformed boosting (98.73$\pm$0.16\% vs. 93.1$\pm$0.5\%) (Fig. \ref{fig_bioi}\textbf{A}) \cite{Eulenberg081364}.
DNN, trained using genetic algorithm with AUC as cost function, performed label free cell classification at higher accuracy (95.5$\pm$0.9\%) than SVM with Gaussian kernel (94.4$\pm$2.1\%), LR (93.5$\pm$0.9\%), NB (93.4$\pm$1\%), and DNN trained with cross entropy (88.7$\pm$1.6\%) (Fig. \ref{fig_bioi}\textbf{A}) \cite{chen2016}.

Colon histopathology images were classified with higher accuracy using DNN and Multiple Instance Learning (97.44\%) compared to K-Means clustering (89.43\%) \cite{xu_deep_2014} (Fig. \ref{fig_bioi}\textbf{B}).
Deep max-pooling CNN detected mitosis in breast histology images with higher accuracy (88\%) in comparison to statistical feature based classification (70\%) (Fig. \ref{fig_bioi}\textbf{B}) \cite{Ciresan2013}. 
Using StSAE with Softmax classifier (SMC), nuclei were more accurately detected from breast cancer histopathology images (88.8 $\pm$ 2.7\%) when compared to: other techniques with SMC-- CNN (88.3 $\pm$ 2.7\%), 3-layer SAE (88.3 $\pm$ 1.9\%), StAE (83.7 $\pm$ 1.9\%), SAE (84.5 $\pm$ 3.8\%), AE (83.5 $\pm$ 3.3\%); SMC alone (78.1 $\pm$ 4\%); and EM (66.4 $\pm$ 4\%) (Fig. \ref{fig_bioi}\textbf{B}) \cite{DBLP0005XLGWTM16}. 

\begin{figure}[!tbh]
	\hspace{-4.5cm} 
    \includegraphics{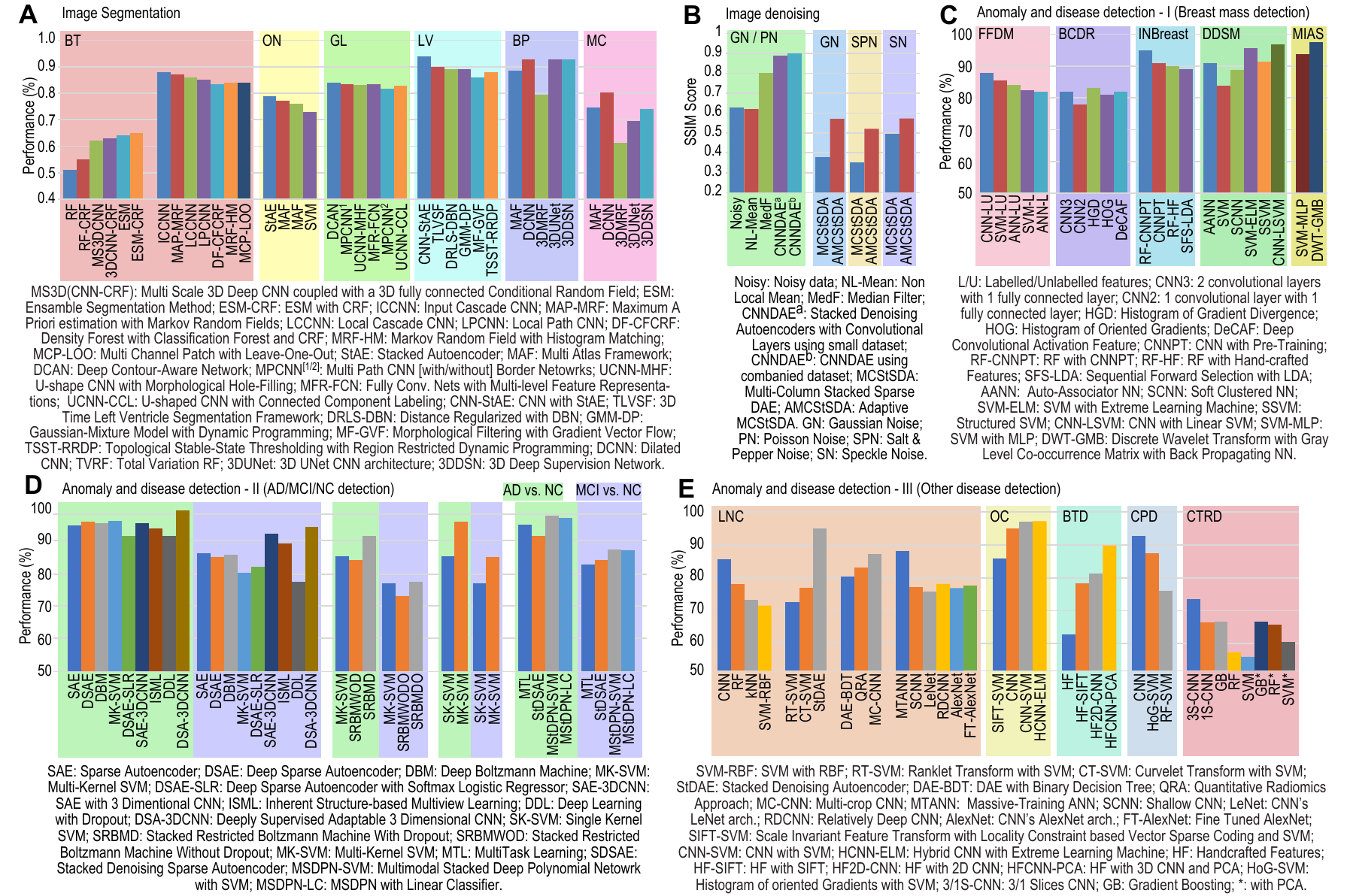}
  \caption{
Performance comparison of representative DL techniques when applied to Medical Imaging. \textbf{A}. Performance of Image Segmentation techniques in segmenting tumors (BT: Brain Tumor), and different organ parts (ON: Optic Nerve, GL: Gland, LV: Left Ventricle of heart, BP: Blood Pool, and MC: Myocardium). \textbf{B}. Image denoising techniques to improve image quality during the presence of Gaussian, Poisson, Salt \& Pepper, and Speckle noise. \textbf{C}. Detecting anomalies and diseases in mammograms. \textbf{D}. Classification and detection of Alzheimer's Disease (AD), Mild Cognitive Impairment (MCI), along with healthy controls (NC). \textbf{E}. Performance of prominent techniques for- Lung Nodule Classification (LNC), Organ Classification (OC), Brain Tumor Detection (BTD), Colon Polyp Detection (CPD), and Chemotherapy Response Detection (CTRD). }
\label{fig_medi}
\end{figure}

\subsection{Medical Imaging}
\label{subsec-medi-perf}
Comparing test results on performance of various DL/non-DL techniques in segmenting medical images to detect pathology or organ parts is reported in Fig. \ref{fig_medi}\textbf{A}. A multi-scale, dual pathway, 11-layers, 3D CNN based method with Conditional Random Fields outperformed RF method (DSC metric values: 63.0 $\pm$ 16.3 vs. 54.8 $\pm$ 18.5) when segmenting brain lesion in MRIs obtained from TBI database. The classifier's accuracy improved when 3 similar networks were ensembled (i.e., Ensembled Method, ESM) and their outputs were averaged (64.5 $\pm$ 16.3 vs. 63.0 $\pm$ 16.3) \cite{kamnitsas_3dcnn_2017}. In a similar task a [two pathway/cascaded] CNN trained using a two-phase training procedure, with local and global features, outperformed other methods participated at the MCCAI-BRATS2013\footnote{See \cite{menze_brats_2015} for MICCAI-BRATS2013.} as reported using Dice coefficients (InputCascadeCNN: 0.88 vs. Tustison: 0.87) \cite{DBLPHavaeiGLJ16}. StAE based method performed similarly or superior to other non-DL methods (see `ON' in Fig. \ref{fig_medi}\textbf{A}, DSC values-- StAE: 0.79 vs. [MAF: 0.77, MAF: 0.76, and SVM: 0.73]) in segmenting optic nerve from MRI data \cite{7420737}. Several DL methods were evaluated in identifying glands in colon histology images, and a DCAN based method outperformed other CNN based methods at the GlaS contest\footnote{See \cite{sirinukunwattana_gl_2017} for MICCAI15-GlaS.} (DCAN: 0.839 vs. [MPCNN$^\texttt{1}$: 0.834, UCNN-MHF: 0.831, MFR-FCN: 0.833, MPCNN$^\texttt{2}$: 0.819, UCNN-CCL: 0.829]) \cite{sirinukunwattana_gl_2017}. Also, left ventricles were segmented from cardiac MRI, where a CNN-StAE based method outperformed other methods (CNN-StAE: 0.94 vs. [TLVSF: 0.9, DRLS-DBN: 0.89, GMM-DP: 0.89, MF-GVF: 0.86, and TSST-RRDP: 0.88]) \cite{Avendi2016}. In segmenting volumetric medical images for blood pool (BP) and myocardium (MC), CNN based methods outperformed other methods as reported using Dice coefficients-- BP (MAF: 0.88, DCNN: 0.93, 3DMRF: 0.87, TVRF: 0.79, 3DUNet: 0.926, 3DDSN: 0.928); and MC (MAF: 0.75, DCNN: 0.8, 3DMRF: 0.61, TVRF: 0.5, 3DUNet: 0.69, 3DDSN: 0.74) \cite{dou_heart_2017}.

DL based methods outperformed other methods in denoising MMM \& dental radiographs \cite{DBLPGondara16}, and brain CT scans \cite{NIPS2013_5030} (Fig. \ref{fig_medi}\textbf{B}). StDAE-CNN performed more accurate denoising in the presence of Gaussian/ Poisson noise (GN/PN) as reported using SSIM scores (Noisy: 0.63, NL Means: 0.62, MedFilt: 0.8, CNNDAE$^a$: 0.89, CNNDAE$^b$: 0.9) \cite{DBLPGondara16}. Adaptive MC-StSDA outperformed MC-StSDA in denoising CT images as reported using PSNR values for GN, Salt \& Pepper (SPN), and Speckle noise (SN) (SSIM\footnote{SSIM $= ($PSNR$-15.0865)/20.069$, with $\sigma_{\texttt{fg}}=10^2$ \cite{hore_psnr_ssim_2010}.}  scores-- GN: 26.5 vs. 22.7; SPN: 25.5 vs. 22.1; SN: 26.6 vs. 25.1) \cite{NIPS2013_5030}.

CNN based methods performed very well in detecting breast mass and lesion in MMM obtained from different datasets (see Fig. \ref{fig_medi}\textbf{C}). MMM obtained from the FFDM database, trained with Labeled and Unlabeled features, CNN outperformed other methods (CNN-LU: 87.9\%, SVM-LU: 85.4\%, ANN-LU: 84\%, SVM-L: 82.5\%, ANN-L: 81.9\%) \cite{sun_cnn_brst_2017}. In detecting masses in MMM from BCDR database, CNN
with 2 convolution layers and 1 fully connected layer (CNN3) performed similar to other methods (CNN3:82\%, HGD: 83\%, HOG: 81\%, DeCAF: 82\%), and CNN with 1 convolution layer and 1 fully connected layer performed poorly (CNN2: 78\%) \cite{arevalo_cnn_brst_2016}. Pre-trained CNN with RF outperformed other methods (e.g., RF with handcrafted features and sequential forward selection with LDA) while analyzing MMM from INBreast database (RF-CNNPT: 95\%, CNNPT: 91\%, RF-HF: 90\%, SFS-LDA: 89\%) \cite{dhungel_dl_brst_2017}. In yet another study, CNN with linear SVM outperformed other methods on MMM from DDSM database (CNN-LSVM: 96.7\%, SVM-ELM: 95.7\%, SSVM: 91.4\%, AANN: 91\%, SCNN: 88.8\%, SVM: 83.9\%) \cite{jiao_cnn_brst_2016}. However, for the MMM form MIAS database, a DWT with back-propagating NN outperformed its SVM/ CNN counterparts (DWT-GMB: 97.4\% vs. SVM-MLP: 93.8\%) \cite{jiao_cnn_brst_2016}.

Despite having been all applied on images from ADNI database, reported methods displayed different performances in detecting and classifying `AD vs. NC' and `MCI vs. NC' (AD and MCI, in short) varied greatly (see Fig. \ref{fig_medi}\textbf{D}). An approach employing deep-supervised-adaptable 3D-CNN (DSA-3D-CNN) outperformed other DL and non-DL methods, as reported using their accuracies, in detecting AD and MCI (AD, MCI-- DSA-3DCNN: 99.3\%, 94.2\% vs. [DSAE: 95.9\%, 85.0\%; DBM: 95.4\%, 85.7\%; SAE-3DCNN: 95.3\%, 92.1\%; SAE: 94.7\%, 86.3\%; DSAE-SLR: 91.4\%, 82.1\%; MK-SVM: 96.0\%, 80.3\%; ISML: 93.8\%, 89.1\%; DDL: 91.4\%, 77.4\%]) 
\cite{DBLPHosseiniAslGE16}. A Stacked RBM with dropout based method outperformed the same method without dropout and multi-kernel based SVM method in detecting AD and MCI (AD, MCI-- SRBMDO: 91.4\%, 77.4\% vs. [SRBMWODO: 84.2\%, 73.1\%; and MK-SVM: 85.3\%, 76.9\%]) 
\cite{Li2014}. In another method with StAE extracted features, MK-SVM was more accurate than SK-SVM method ([AD, MCI]-- MK-SVM: [85.3\%, 76.9\%] vs. SK-SVM: [95.9\%, 85.0\%]) 
\cite{suk_da_2013}. Another method, where features from MRI and PET were fused and learned using multi-modal stacked Deep Polynomial Network (MStDPN) algorithm, outperformed other multimodal learning methods in detecting AD and MCI (AD, MCI-- MTL: 95.38\%, 82.99\%; StSDAE: 91.95\%, 83.72\%; MStDPN-SVM: 97.13\%, 87.24\%; MStDPN-LC: 96.93\%, 86.99\%) 
\cite{shi_sdpn_ad_2017}.

Different techniques reported varying accuracies in detecting a range of anomalies from different medical images (Fig. \ref{fig_medi}\textbf{E}). 
CNN had better accuracy in classifying ILD (85.61\%) when compared to RF (78.09\%), kNN (73.33\%), and SVM-RBF (71.52\%) \cite{7422082}. LNC were accurately done using StDAE (95\%) compared to RT-SVM (72.5\%) and CT-SVM (77\%) \cite{chengchen2016}. Multi-crop CNN achieved better accuracy (87.14\%) than DAE with binary DT (80.29\%) and quantitative radiomics based approach (83.21\%) in LNC \cite{shen_cnn_ln_2017}. MTANNs outperformed CNNs variants in LNC (MTANN: 88.06\% vs. [SCNN: 77.09\%, LeNet: 75.86\%, RDCNN: 78.13\%, AlexNet: 76.85\%, FT-AlexNet: 77.55\%]) \cite{tajbakhsh_cnn_ln_2017}. Hierarchical CNN with ELM outperformed other CNN and SVM methods (HCNN-NELM: 97.23\% vs. [SIFT-SVM: 89.79\%, CNN: 95\%, CNN-SVM: 97.05\%]) in classifying digestive organs \cite{7419037}. Multi-channel CNN with PCA and handcrafted features better detected BT (89.85\%) in comparison to 2D-CNN (81.25\%), scale-invariant transform (78.35\%), and manual classification with handcrafted features (62.8\%) \cite{Nie2016}. A 2D CNN based method trained with stochastic gradient descent learning outperformed other non-DL methods (AUC values-- CNN: 0.93 vs. [HOG-SVM: 0.87, and RF-SVM: 0.76]) in detecting colon polyp from CT colonographs \cite{DBLPRothLLYSKKS15}. In addition, 3Slice-CNN was successfully employed to detect Chemotherapy response in PET images which outperformed other shallow methods (3S-CNN: 73.4 $\pm$ 5.3\%, 1S-CNN: 66.4 $\pm$ 5.9\%, GB: 66.7 $\pm$ 5.2\%, RF: 57.3 $\pm$ 7.8\%, SVM: 55.9 $\pm$ 8.1\%, GB-PCA: 66.7 $\pm$ 6.0\%, RF-PCA: 65.7 $\pm$ 5.6\%, SVM-PCA: 60.5 $\pm$ 8.0\%) \cite{Ypsilantis7036}.

\subsection{[Brain/Body]-Machine Interfaces}
\label{subsec-bmi-perf}
\begin{figure}[!htbp]
\hfill
\includegraphics{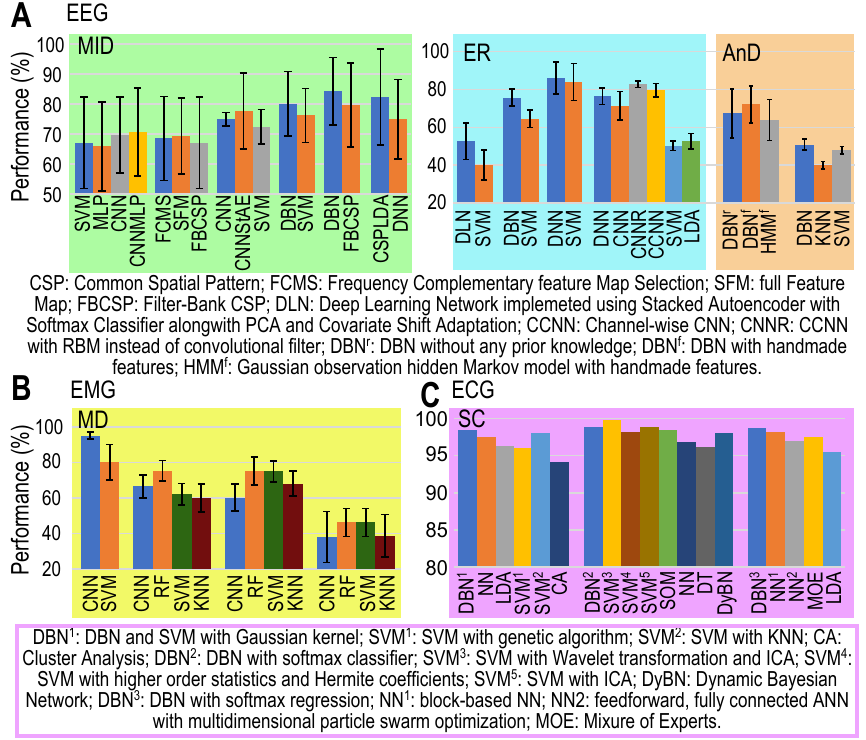}
\caption{
Accuracy comparison of DL and conventional ML techniques when applied to BMI signals. 
\textbf{A}. Performance comparison in detecting motor imagery (MID), recognizing emotion and cognitive states (ER), and detecting anomaly (AnD) from EEG signals. \textbf{B}. Accuracies of movement decoding (MD) from EMG signals. \textbf{C}. Accuracies of ECG signal classification (SC).}
\label{fig_bmi}
\end{figure}

Test results in the form of performance comparison of DL/non-DL methods applied to EEG data to detect MoI, emotion \& affective state, and anomaly are shown in Fig. \ref{fig_bmi}\textbf{A}.

A linear parallel CNN with MLP classified EEG energy dynamics more accurately (70.6\%), than SVM (67.0\%), MLP (65.8\%), and CNN (69.6\%) to detect MoI from BCI competition (BCIC) IV-2a dataset \cite{sakhavi_mi_2015}. 
CNN better classified FCMS of augmented CSP and SFM as features (68.5\% and 69.3\%) than filter-bank CSP (67.0\%) to detect MoI from BCIC IV-2a dataset \cite{yang2015}.
CNN, StAE, and their combination (CNN-StAE) were tested in classifying MoI from BCIC IV-2b EEG data. Using time, frequency \& location information as features, CNN-StAE achieved best accuracy (77.6 $\pm$ 2.1\%) in comparison to SVM (72.4 $\pm$ 5.7\%), CNN (74.8 $\pm$ 2.3\%) and StAE (57.7 $\pm$ 5.5\%) \cite{tabar_cnn_mi_eeg_2017}. 
A DBN with Ada-boost based classifier had higher accuracy ($\sim$81\%) than SVM ($\sim$76\%) in classifying hand movements from EEG \cite{An2014}.
Another DBN based method reported better accuracy (0.84) using frequency representations of EEG (using FFT and wavelet package decomposition) rather than FCSP (0.8), and CSP (0.76) in classifying MoI \cite{lu_rbm_mi_2016}.
A DNN based method, with layerwise relevance propagation heatmaps, performed comparable (75\%) to CSP-LDA (82\%) in MoI classification \cite{DBLPSturmBSM16}.

A DLN was built using StAE with PCA and covariate shift adaptation to classify valence and arousal states from DEAP EEG data with multichannel PSD as features. The mean accuracy of the DLN was 52.7 $\pm$ 9.7\% compared to SVM (40.1 $\pm$ 7.9\%) \cite{jirayucharoensak2014}.
A supervised DBN based method classified affective states more accurately (75.6 $\pm$ 4.5\%) when compared to SVM (64.7 $\pm$ 4.6\%)  by extracting deep features from thousands of low level features using DEAP EEG data \cite{li_dbn_as_2013}. 
A DBN based method, with differential entropy as features, explored critical frequency bands and channels in EEG, and classified three emotions (positive, neutral, and negative) with higher accuracy (86.1\%) than SVM (83.9\%) \cite{zheng2015investigating}.
As reported through Az-score, in predicting driver's drowsy and alert state from EEG data, CCNN and CNNR methods outperformed (79.6\% and 82.8\% respectively) other DL (CNN: 71.4$\pm$7.5\% and DNN: 76.5$\pm$4.4\%) and non-DL (LDA: 52.8\%$\pm$4, and SVM: 50.4$\pm$2.5\%) methods \cite{7383844}.

DBN was used to model, classify, and detect anomalies from EEG waveforms. It has been reported that using raw data a comparable classification and superior anomaly detection accuracy (50 $\pm$ 3\%) can be achieved 
compared to SVM (48$\pm$2\%), and KNN (40$\pm$2\%) classifiers \cite{wulsin2011}.
Another DBN and HMM based method performed comparable sleep stage classification from raw EEG data (67.4 $\pm$ 12.9\%) with respect to DBN with HMM and features (72.2 $\pm$ 9.7\%), and Gaussian observation HMM with features (63.9 $\pm$ 10.8\%) \cite{langkvist2012}.

Fig. \ref{fig_bmi}\textbf{B} shows performances of various methods in decoding movements (MD) from (s)EMG. 
A CNN based method's hand movement classification accuracy using three sEMG datasets (from the Ninapro database) were comparable to other methods (CNN vs. [kNN, SVM, RF]) -- Dataset1: 66.6 $\pm$ 6.4\% vs. [60 $\pm$ 8\%, 62.1 $\pm$ 6.1\%, 75.3 $\pm$ 5.7\%]; Dataset2: 60.3 $\pm$ 7.7\% vs. [68 $\pm$ 7\%, 75 $\pm$ 5.8\%, 75.3 $\pm$ 7.8\%]; and Dataset3: 38.1 $\pm$ 14.3\% vs. [38.8 $\pm$ 11.9\%, 46.3 $\pm$ 7.9\%, 46.3 $\pm$ 7.9\%] \cite{DBLPAtzoriCM16}.
Another method, a user-adaptive one, using CNN with deep feature learning, decoded movements more accurately compared to SVM (95 $\pm$ 2\% vs. 80 $\pm$ 10\%) \cite{7457459}. 

Fig. \ref{fig_bmi}\textbf{C} compares different techniques' performances in classifying ECG signals from MIT-BIH arrhythmia database and detecting anomalies in them.
A nonlinear SVM with Gaussian kernel (DBN$^\texttt{1}$) outperformed (98.5\%) NN (97.5\%), LDA (96.2\%), SVM with genetic algorithm (SVM$^\texttt{1}$: 96.0\%), SVM with kNN (SVM$^\texttt{2}$: 98.0\%), Wavelet with PSO (88.8\%), and CA (94.3\%) in classifying ECG features extracted using DBN \cite{7023547}.
Comparable accuracy in classifying ECG beats were obtained using DBN with softmax (DBN$^\texttt{2}$: 98.8\%) compared to SVM with Wavelet and ICA (SVM$^\texttt{3}$: 99.7\%), SVM with higher order statistics and Hermite coefficients (SVM$^\texttt{4}$: 98.1\%), SVM with ICA (SVM$^\texttt{5}$: 98.8\%), DT (96.1\%), and Dynamic Bayesian network (DyBN: 98\%) \cite{DBLPYanQWZ0W15}.
Using DBN (with contrastive divergence and persistent contrastive divergence learning), arrhythmias were classified more accurately (98.6\%) in comparison to block NN$^\texttt{1}$ (98.1\%), feed-forward based NN with PSO (NN$^\texttt{2}$: 97.0\%), mixture of experts (97.6\%), and LDA (95.5\%) \cite{wu_ecg_dbn_2016}.

\section{Open Issues and Future Perspectives}
\label{sec-issues-persp}
Overall, it is believed that brain solves problems through reinforcement learning and neuronal networks organized as hierarchical processing systems. Though since the 1950's the field of AI has been trying to adopt and implement this strategy in computers, notable progress has been seen only recently due to our better understanding about learning systems, increase of computational power, decline of computing costs, and last but not the least, the seamless integration of different technological and technical breakthroughs. However, there are still situations where these methods fail, underperforming traditional methods, and, therefore, must be improved. Below we outline, what in our opinion are, the shortcomings of current techniques, the existing open research challenges, and speculate about some future perspectives that will facilitate further development and advancement of the field.

The combined computational capability and flexibility provided by the two prominent ML methods (i.e., DL and RL)  has also limitations \cite{ravi_dl_2017}. Both of these methods require heavy computing power and memory and, therefore, they are not worthy of being applied to moderate size datasets. Additionally, the theory of DL is not completely understood, making the high level outcomes obscure and difficult to interpret. This turns into a situation when the models are considered as `Black box' \cite{Erhan-vis-techreport-2010}. In addition, like other ML techniques, DL is also susceptible to misclassification \cite{nguyen_dl_fool_2015} and overclassification \cite{szegedy_ipnn_2014}. Furthermore, in representing action-value pairs in RL, it is not possible to use all nonlinear approximators which may cause instability or even divergence in some cases \cite{mnih_human-level_2015}. Also, bootstrapping makes many of the RL algorithms NP hard and inapplicable to real-time applications as they are too slow to converge and in some cases too dangerous (e.g., autonomous driving). Moreover, very few of the existing techniques support harnessing the potential power of distributed and parallel computation through cloud computing. Arguably, in case of cloud, distributed, and parallel computing, data privacy and security concerns are still prevailing \cite{mahmud_soa_2012}, and real-time processing capability of the gigantic amount of experimentally acquired data is still underdeveloped \cite{mahmud_qspike_2014,mahmud_webqst_2014}.

To proceed towards mitigating the shortcomings and addressing the open issues, first of all, improving the existing theoretical foundations of DL on the basis of experimental data becomes crucial to be able to quantify the performances of individual NN models \cite{angelov_dl_challenges_2016}. These improvements should be able to address issues like-- specific assessment of an individual model's computational complexity and learning efficiency in relation to well defined parameter tuning strategies, the ability to generalize and topologically self-organize based on data-driven properties. Also, novel data visualization techniques should be incorporated so that the interpretation of data becomes intuitive and less cumbersome. 
In terms of learning strategies, updated hybrid on- and off-policy with new advances in optimization techniques are required. The problems pertaining to observability of RL are yet to be completely solved, and optimal action selection is still a huge challenge.

As seen in Table \ref{tab:main}, there are great opportunities to employ deep RL in Biological data mining. For example, deriving dynamic information from Biological data coming from multiple levels to reduce data redundancy and discover novel biomarkers for disease detection and prevention. Also, new unsupervised learning for deep RL methods are required to shrink the necessity of large-set of labeled data at the training phase. Multitasking and multiagent learning paradigm should advance in order to cope with dynamically changing problems. 

In addition, to keep up with the rapid pace of data growth in the biological application domains, computational infrastructures in terms of distributed and parallel computing tailored to those applications are needed.

\section{Conclusion}
\label{sec-conclusion}
The recent bliss of technological advancement in Life Sciences came with the huge challenge of mining the multimodal, multidimentional and complex Biological data. Triggered by that call, interdisciplinary approaches have resulted in development of cutting edge machine learning based analytical tools. The success stories of artificial neural networks, deep architectures, and reinforcement learning in making machines intelligent are well known. Furthermore, computational costs have dropped, computing power has surged, and \textit{quasi}unlimited solid-state storage is available at reasonable price. These factors have allowed to combine these learning techniques to reshape machines' capabilities to understand and decipher complex patterns from Biological data. To facilitate wider deployment of such techniques and to serve as a reference point for the community, this article provides-- a comprehensive survey of the literature of techniques' usability with different Biological data; a comparative study on performances of various DL techniques, when applied to the data coming from different application domains, as reported in the literature; and highlights of some open issues and future perspectives.

\section*{Acknowledgment}
The authors would like to thank Dr. Pawel Raif and Dr. Kamal Abu-Hassan for useful discussions during the early stage of the work. This work was supported by the ACSLab (\href{http://www.acslab.info}{www.acslab.info}).



\nolinenumbers

\bibliographystyle{IEEEtran}

\bibliography{ref}

\end{document}